\documentclass[conference]{IEEEtran}
\IEEEoverridecommandlockouts
\usepackage{cite}
\usepackage{amsmath,amssymb,amsfonts}
\usepackage{algorithmic}
\usepackage{graphicx}
\usepackage{textcomp}
\usepackage{xcolor}

\usepackage{subcaption}
\usepackage[utf8]{inputenc} 
\usepackage[T1]{fontenc}    
\usepackage{hyperref}       
\hypersetup{
    colorlinks = true,
    linkcolor=red,
    anchorcolor=black,
    citecolor=red,
    urlcolor=magenta
}

\usepackage{url}            
\usepackage{booktabs}       
\usepackage{amsfonts}       
\usepackage{nicefrac}       
\usepackage{microtype}      
\usepackage{xcolor}
\usepackage{amsmath}
\usepackage{mathtools}
\usepackage{amsthm}
\usepackage{graphicx}
\usepackage{caption}
\usepackage{lineno}
\usepackage{comment}
\usepackage{textcomp}
\usepackage{multirow}

\usepackage{lipsum}
\usepackage{tikz}
\usepackage{xcolor}
\usepackage{setspace}
\usepackage[flushleft]{threeparttable}

\usepackage{bbding}
\usepackage{pifont}
\usepackage{amssymb}
\usepackage{pifont}

\newcommand*\circled[1]{\tikz[baseline=(char.base)]{
            \node[circle,fill=.,inner sep=0.8pt] (char) {\textcolor{white}{#1}};}}
        
\newcommand*{\Mname}{DSXplore}
\newcommand*{\affaddr}[1]{#1} 

\newcommand*{\email}[1]{\texttt{#1}}

\usepackage[ruled,linesnumbered]{algorithm2e}
\DeclarePairedDelimiterX\set[1]\lbrace\rbrace{#1}

\def\BibTeX{{\rm B\kern-.05em{\sc i\kern-.025em b}\kern-.08em
    T\kern-.1667em\lower.7ex\hbox{E}\kern-.125emX}}
\begin{document}

\title{DSXplore: Optimizing Convolutional Neural Networks via Sliding-Channel Convolutions
\thanks{Paper accepted at IPDPS-2021.}
}

\author{
Yuke Wang, Boyuan Feng, and Yufei Ding\\  
\affaddr{Department of Computer Science,}\\
\affaddr{University of California, Santa Barbara.} \\
\email{\{yuke\_wang,boyuan,yufeiding\}@cs.ucsb.edu} \\ 
}

\maketitle
{
\setstretch{1}
\begin{abstract}
As the key advancement of the convolutional neural networks (CNNs), depthwise separable convolutions (DSCs) are becoming one of the most popular techniques to reduce the computations and parameters size of CNNs meanwhile maintaining the model accuracy. It also brings profound impact to improve the applicability of the compute- and memory-intensive CNNs to a broad range of applications, such as mobile devices, which are generally short of computation power and memory. However, previous research in DSCs are largely focusing on compositing the limited existing DSC designs, thus, missing the opportunities to explore more potential designs that can achieve better accuracy and higher computation/parameter reduction. Besides, the off-the-shelf convolution implementations offer limited computing schemes, therefore, lacking support for DSCs with different convolution patterns.

To this end, we introduce, DSXplore, the first optimized design for exploring DSCs on CNNs.
Specifically, at the algorithm level, DSXplore incorporates a novel factorized kernel -- sliding-channel convolution (SCC), featured with input-channel overlapping to balance the accuracy performance and the reduction of computation and memory cost. 
SCC also offers enormous space for design exploration by introducing adjustable kernel parameters.
Further, at the implementation level, we carry out an optimized GPU-implementation tailored for SCC by leveraging several key techniques, such as the input-centric backward design and the channel-cyclic optimization.
Intensive experiments on different datasets across mainstream CNNs show the advantages of DSXplore in balancing accuracy and computation/parameter reduction over the standard convolution and the existing DSCs. 

\end{abstract}
\section{introduction}
With the increasing popularity of the AI-driven edge computing and internet of things (IoTs), convolutional neural networks (CNNs) have entered a new era with the plethora of tiny devices, which have limited resource, such as the power, and memory budget. This makes the CNNs that are small in parameter size with low computation costs (FLOPs) highly in demand. Among numerous research and industry efforts, depthwise separable convolutions (DSCs) attract a lot of attentions, largely because of their stunning success in reducing FLOPs and parameters. 

Existing works around DSCs have been widely studied mainly from algorithmic perspective. MobileNet~\cite{mobilenet_2017_howard} has been proposed by replacing the standard convolution with the depthwise separable convolution (depthwise (DW) + pointwise (PW) convolution) to reduce the model parameters and computation cost significantly. 
Inspired by the success of MobileNet, Xception~\cite{xception} architecture has been built to largely simplify more complicated CNNs, such as Inception~\cite{inceptionV4} (with large number of layers and residual connections) by leveraging DSCs. 
The major assumption of these works is that the cross-channel correlations and the spatial correlations can be effectively decoupled and there is no need to map them at the same time. Therefore, the high-cost standard convolution can be effectively divided into the lightweight DW convolution to capture the spatial information and PW convolution to capture the channel-wise information.

However, these existing efforts on DSCs are still initial, since they overlook some key points that could be potentially leveraged for further parameter and computation reduction. And we believe there are several reasons behind. 

\textbf{First, there are limited DSC designs to balance accuracy performance and the size of computation/parameters}. 
Existing work on DSCs is mostly derived from the DW+PW design for standard convolution replacement. 
The more effective DSC schemes that can potentially deliver better accuracy and model size trade-offs still remain uncovered. 
For example, we can further reduce the computation cost and parameter size by combining the group convolution (GC) (dividing the input and output channel into the same number of groups and only applying standard convolution within each group) with PW. 

\textbf{Second, there is a lack of efficient implementation support for new factorized kernels}.
Existing works on DSCs heavily rely on the deep-learning infrastructure with standard/group convolutions for their factorized kernel implementation. 
For example, the DW convolution can be expressed as the extreme case of the GC with the number of groups equal the input channels, while the PW convolution can be expressed as another special case of standard convolution with the $1\times1$ kernel spatial dimension. 
Therefore, the better factorized kernel that may bring better accuracy and lower computation and memory costs but not in the above categories cannot leverage the existing convolutional primitives for an effective implementation.

To this end, we propose, \Mname, the first optimized design for exploring the long-existing ``buried'' DSC potentials. The highlight of \Mname~is our novel factorized convolution kernel -- sliding-channel convolution (SCC), which can effectively reduce the computation and parameter size meanwhile maintaining model accuracy to a great extent. In contrast to the previous fixed DW+PW in most existing DSC designs, \Mname~provides an enormous space for exploring new DSC designs by introducing a set of adjustable parameters to SCC -- the number of channel groups ($cg$) and input-channel overlapping ($co$). Furthermore, we present an optimized implementation of SCC on GPUs by capturing the specialty (\textit{e.g.}, cyclic-channel pattern) of SCC and tailoring parallel computation (\textit{e.g.}, improving thread-level parallelism and reducing atomic operations).
\begin{figure*} [t] \small
    \centering
    \includegraphics[width=\textwidth]{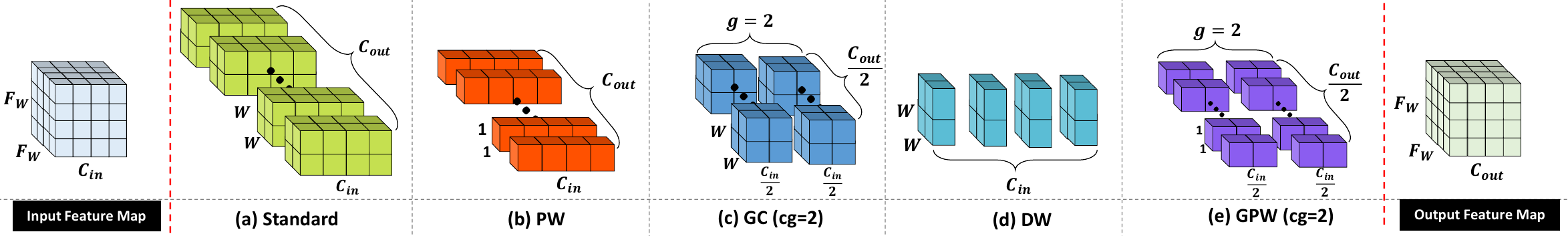}
    \caption{Comparison of input/output feature map before/after convolution and the filter dimension of existing CNN convolutions.}
    \vspace{-10pt}
    \label{fig: Comparison of different convolutional kernels.}
\end{figure*}

Overall, we make the following contributions
\begin{itemize}
    \item We propose the first optimized design to uncover the potentials of DSC. We harness our novel sliding-channel convolution (SCC) to balance the accuracy performance and the reduction of computation and memory cost. Moreover, SCC offers an enormous design exploration space with parameterized design strategy.
    \item We carry out an optimized GPU-implementation tailored for SCC design by incorporating several key techniques, including the output-centric forward and input-centric backward optimization, and the optimization based on the convolutional specialty (cyclic channel) of its filters. 
    \item We integrated our SCC design with the original Pytorch framework as the drop-in replacement of the existing DSCs to facilitate the training and inference of \Mname\footnote{\Mname~is open-sourced at \url{ https://github.com/YukeWang96/DSXplore.git}}~in an end-to-end fashion.
    \item Intensive experiments show \Mname~achieves better accuracy and lower computation/memory cost compared with the existing DSC. And our proposed optimized GPU implementation can overcome the implementation challenges of SCC by providing notable speedup compared with the implementations by compositing the highly-optimized Pytorch operators.
\end{itemize}

\section{Background and Related Work}
\subsection{Standard Convolution}
The standard convolution is the most widely used deep-learning operation in many CNNs~\cite{resnext, vgg, sandler2018mobilenetv2}, which targets on images (\textit{a.k.a.}, feature map). 
In general, we annotate the dimension of 3D feature map as $F_w\times F_w\times C_{in}$, where $F_W$ is the 2D spatial dimension of the feature map while $C_{in}$ is the number of input channels.

The standard convolution relies on a set of ($C_{out}$) standard convolutional filters, each of which has the size of $W\times W\times C_{in}$ parameters to generate output feature maps, as shown in Figure~\ref{fig: Comparison of different convolutional kernels.}(a), where the $W$ is the spatial dimension of the filters (In general, filters usually have a square shape of $W\times W$), $C_{in}$ is the number of channels for the input feature map, and $C_{out}$ is the number filters (equals the channels for the output feature map, since each filter only generates the feature map in one output channel). After applying the standard convolution on the input with the shape of $F_{w}\times F_{w}\times C_{in}$, we will get the output feature map $O$, which has the shape of $F_{w}\times F_{w}\times C_{out}$.
Note that the mainstream CNNs~\cite{resnext,vgg,mobilenet_2017_howard} generally maintain the same feature map spatial dimension at different convolutional layers while only changing the number of the channels across different layers. 

Formally, for standard convolution, we have
\begin{equation}
    O_{m, n, c} = \sum\limits_{i,j,a}K_{i,j,a,c} * F_{m+i-1,n+j-1,a}
\end{equation}
where $O_{m,n,c}$ is one pixel point in the output feature map; $m$ and $n$ are the spatial indexes in the output feature map ($0\leq m< F_w$ and $0\leq n< F_w$); 
$a$ is the channel index in the input feature map ($0\leq a< C_{in}$);
$c$ is the channel index in the output feature map ($0\leq c< C_{out}$); $i$, $j$, and $a$ are the index used to accumulated the elementwise multiplication values between input feature map and one filter, which is essentially a cube with the shape of $K\times K\times C_{in}$.

The standard convolution not only captures the spatial information by iteratively ``gathering'' a $W\times W$ square in a 2D sliding-window fashion in each channel but also effectively fuses the information across different channels, as indicated in the Figure~\ref{fig: Comparison of different convolutional kernels.}(a), where each kernel filter will ``fuse'' the information from all $C_{in}$ channels of input feature maps.

\subsection{Factorized Convolution}
While the standard convolution gains success in CNNs across different applications settings (\textit{e.g.}, computer vision), its high computation and memory complexity significantly make it challenging to get widely adopted on devices with limited capability of computation and memory resources. 

To break this hurdle, a set of factorized kernels and their combinations are introduced as a drop-in replacement for the standard convolution to reduce the computation and memory cost meanwhile maintaining the model prediction power. Existing factorized kernels can be divided into four major categories: 1) Pointwise Convolution (PW)~\cite{inceptionV4}, which is a standard convolution with $1\times1$ spatial size, as shown in Figure~\ref{fig: Comparison of different convolutional kernels.}(b); 2) Group Convolution~(GC)~\cite{alexnet_2012} that divides input channels into several groups and performs standard convolution within each group, as exemplified in Figure~\ref{fig: Comparison of different convolutional kernels.}(c) with two groups; 3) Depthwise Convolution (DW)~\cite{depthwise_origin} which calculates spatial convolution per channel or can be regarded as an extreme case of GC when the group number equals the number of the input channels, as shown in Figure~\ref{fig: Comparison of different convolutional kernels.}(d); 4) Group Pointwise Convolution~(GPW)~\cite{shufflenet_zhang_2017} that further splits PW into groups, as exemplified in Figure~\ref{fig: Comparison of different convolutional kernels.}(e) with two groups.

The most successful example adopted in many CNNs (\textit{e.g.}, Xception~\cite{xception} and MobileNet~\cite{mobilenet_2017_howard}) is the depthwise separable convolution (\textbf{DSC}). It breaks the original standard convolution into two parts: \textbf{depthwise} (DW) convolution and \textbf{pointwise} (PW) convolution. The first step (DW) applies $C_{in}$ different $W\times W \times 1$ filters to each of the $C_{in}$ input channels independently, which can be formalized as Equation~\ref{equ: depth-wise convolution}
\begin{equation} \label{equ: depth-wise convolution}
\hat{O}_{m,n,a} = \sum\limits_{i,j}K_{i,j,a}*F_{m+i-1, n+j-1,a}
\end{equation}
The second step (PW) applies a filter with $1\times1$ spatial dimension. 
As shown in Equation~\ref{equ: point-wise convolution}.
\begin{equation} \label{equ: point-wise convolution}
O_{m,n,c} = \sum\limits_{a}K_{a,c}*F_{m-1,n-1,a}
\end{equation}
Compared with the standard convolution, DSC brings two folds of benefits. First, it largely reduces the size of weight parameters. In the standard convolution, we have $W\times W\times C_{in} \times C_{out}$ parameters, while in DSC, we only have $W\times W\times C_{in} +  C_{in}\times C_{out}$ parameters, which is only $\frac{1}{C_{out}} + \frac{1}{W^2}$ of the parameters in the standard convolution. Second, it can largely reduce the total number of computations (FLOPs) compared with the standard convolution kernel. Specifically, the standard convolution requires $F_{w}\times F_{w}\times C_{out} \times W\times W\times C_{in}$ operations, whereas DSC only requires $C_{in}\times F_{w}\times F_{w} \times W\times W +  C_{out} \times F_{w} \times F_{w} \times C_{in}$, which is $\frac{1}{C_{out}} + \frac{1}{W^2}$ of the number of operations in the standard convolution. 
Further study, such as Shift Convolution~\cite{shift-convolution}, adapts the depthwise convolution by using a shift matrix for a more specialized spatial-wise information fusion at individual input channels. 

However, the underlying reason why these designs can be successful is not clear. 
Our work, in contrast, introduces a brand-new DSC convolution kernel -- sliding-channel convolution (SCC), focusing on the channel-wise information fusion. 
It offers explainable parameters (the number of channel groups and the input-channel overlapping ratio) to effectively capture cross-channel information with more flexibility.

\subsection{Sparse Convolution}
As an another way of reducing computation and compressing model size, sparse convolution~\cite{patDNN, Scalpel, weightprunningEccv2018, ma2020pconv, AutoCompress} has been proposed and widely studied. 
Depending on the granularity of the pruning, existing pruning strategies can also be categorized into two major types, non-structured and structured pruning. The non-structured pruning aims to maximize the ratio of reducing the parameters and saving FLOPs. However, it fails to consider the implementation complexity on modern hardware due to its computation irregularity after pruning.
The structured pruning overcomes the weakness of non-structured pruning through applying coarse-grained pruning strategy to largely maintain the computation regularity. 
However, it may compromise model accuracy to some extent due to such an  ``imprecise'' nature of the structural pruning.

Furthermore, the modern deep-learning frameworks, such as Pytorch~\cite{pytorch} and Tensorflow~\cite{abadi2016tensorflow}, still lack of efficient programming library support for these pruning operations (\textit{a.k.a.}, sparse kernels) on CPUs and GPUs. And most of these sparse kernels are only evaluated in their accuracy performance by masking the feature maps or weight parameter with zero values. At the same time, some of the pruning method is highly input-sensitive, which may lead to non-deterministic model accuracy when encountering the ``unseen'' real-world datasets. 
In contrast, our work focusing on factorized kernel design is essentially orthogonal to these pruning approaches but share the commonality of saving computation and parameters of CNNs. And the difference is that our work is to redesign the overall kernel filters and their operation patterns (\textit{e.g.}, the input-to-output channel mapping of kernel filters), while these pruning works target at modifying individual filters or feature maps, such as masking out specific tensors based on their values. We also believe that factorize kernel + pruning is a potential research direction, but it is not our current focus in this paper. 

\section{Sliding-Channel Convolution}

Motivated by previous research on DSCs, we believe some ``hidden'' dimensions yet to be explored in this direction. 
And there are three major questions to be answered: 

1) Is it possible to further save the parameters and the computation costs by crafting a new DSC convolution scheme? 

2) How could we maintain the model accuracy performance by applying such a design?

3) How could we implement such a new design to facilitate the end-to-end training on modern hardware, like GPUs?

\subsection{Sliding-Channel Convolution}
To this end, we propose a brand new convolution scheme, named sliding-channel convolution (SCC), 
in place of the PW convolution in the DW+PW design to capture cross-channel information more effectively. 
Specifically, it combines the GC to reduce the parameter and computations in the original PW convolution. Most importantly, it overlaps the input-channels of each filter to ``recover'' the information that is ignored by applying GC. 
Such specialty makes our SCC different from previous designs (PW and GPW) in different perspectives.
\begin{figure} [t] \small
    \centering
    \includegraphics[width=0.5\textwidth]{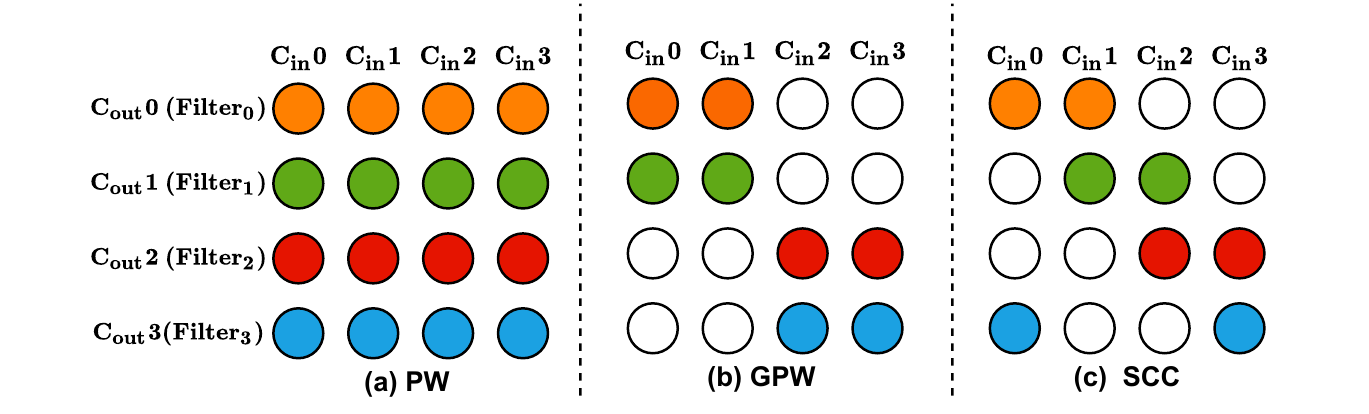}
    \caption{Comparison of our proposed SCC to PW and GPW on input-to-output channel mapping of filters. Note that circles in the same color represent input channels covered by one filter.}
    \label{fig: Comparison of PW, GC and SCC.}
\end{figure}

\begin{table}[h] \small
    \caption{Comparison among SCC, PW, and GPW.}
    \centering
    \begin{threeparttable}
    \vspace{-3pt}
    \begin{tabular}{|c|c c c|}
    \hline
        \textbf{Convolution} & \textbf{FLOPs} & \textbf{Params.} & \textbf{Acc.} \\
    \hline
    PW$^\dagger$ & High & High & High \\
    GPW$^*$ & Low & Low & Low  \\
    \textbf{SCC} & \textbf{Low} & \textbf{Low} & \textbf{High} \\
    \hline
    \end{tabular}
     \begin{tablenotes}
      \footnotesize
      \item $^\dagger$: PW can be seen as SCC with 1 group with 100\%  channel overlapping of adjacent filters.
      \item $^*$: GPW can be seen as SCC with $m$ groups with 0\% channel overlapping of adjacent filters. 
      $m$ is a parameter that can be determined by users.
     \end{tablenotes}
    \end{threeparttable}
    \label{table: comparison with PW and GW.}
    \vspace{-15pt}
\end{table}

\textbf{Compared with PW:} As the second step in DW+PW design, the major role of PW is to cross-channel information by a standard convolution with $1\times1$ filter. While SCC also shares a similar goal as PW to unify channel-wise information, each kernel filter of SCC only needs to look through a part of the input channels, whereas each kernel filter of PW has to look through all input channels. As illustrated in Figure~\ref{fig: Comparison of PW, GC and SCC.}(a), the $C_{out}0$ in SCC only need to ``gather'' information from the input channel $C_{in}0$ and $C_{in}1$, whereas $C_{out}0$ in PW requires information from channel $C_{in}0$, $C_{in}1$, $C_{in}2$ and $C_{in}3$. Note that we use the circle to represent the feature map ($F_w\times F_w$) on each channel. The benefits of our SCC design compared with PW can be highlighted by comparing line 3 and line 1 in Table~\ref{table: comparison with PW and GW.}, where SCC reduces the FLOPs and parameters meanwhile maintaining accuracy performance.
\begin{figure*} [t] \small
    \centering
    \includegraphics[width=\textwidth]{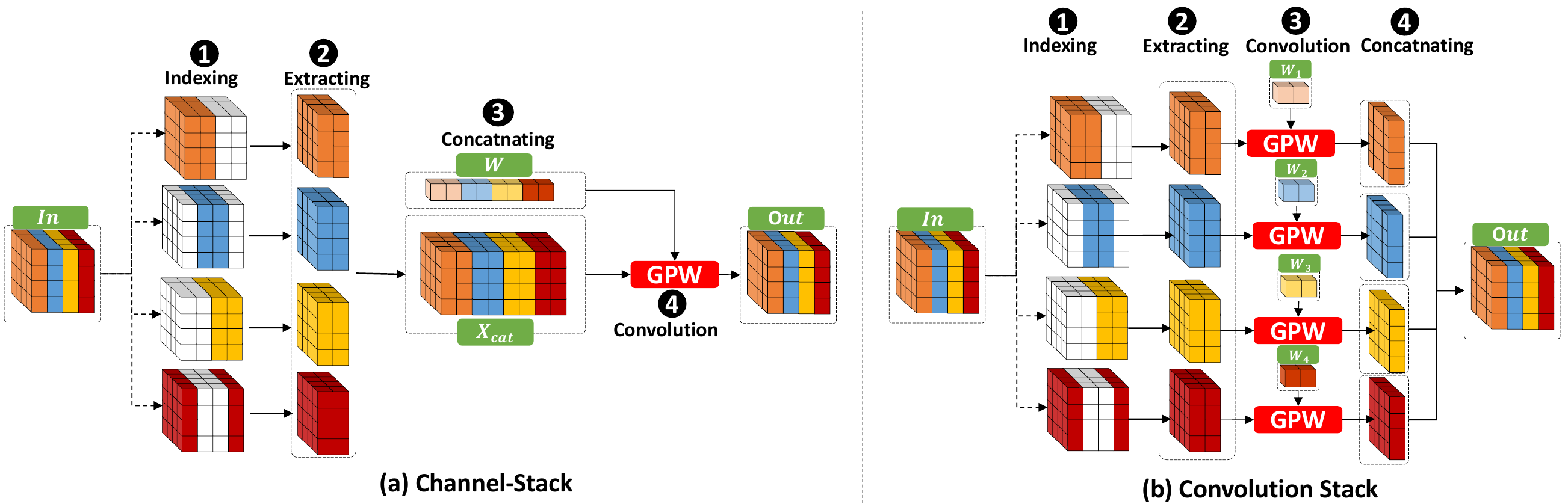}
    \caption{Two proposed Implementations for SCC via Pytorch-operator composition: \textit{channel-stack} and \textit{convolution-stack}.}
    \label{fig: Implementation via Pytorch-operator composition.}
\end{figure*}

\textbf{Compared with GPW:} As the variant of the PW convolution, GPW has been studied to amortize the computation and parameter costs of PW. 
While sharing the commonality of reducing the model complexity as GPW, each filter of our SCC would cover different ``windows'' of input channels that are partially overlapped with their adjacent filters.
The input-channel overlapping in SCC serves as an essential ``bridge'' to communicate the different input channel information that is originally segregated by channel grouping in GPW.
Filters in the conventional GPW, on the other hand, would either fully overlap with their adjacent filters or no overlap with their adjacent filters in terms of input channels they cover. This may lead to 1) redundant information from the same input-channel window, and 2) missing information spanning across different input-channel windows. As illustrated in Figure~\ref{fig: Comparison of PW, GC and SCC.}(b), the first half of the output channels only uses the first half of the input channels from the input feature map, while the second half of the output channels only uses the second half of the input channels. In contrast, in our SCC design (Figure~\ref{fig: Comparison of PW, GC and SCC.}(c), the adjacent two filters would never occupy the same part of input-channel window. For example, filter 2 overlaps with the filter 1, thus, extracting information from both $C_{in}1$ and $C_{in}2$, which cannot be captured in GPW. Besides, we introduce the channel-circulation scheme that further strengthens the capability to extract information, \textit{i.e.}, the last input channel is logically connected with the first input channel to form a cycled input channels. Therefore, as shown in Figure~\ref{fig: Comparison of PW, GC and SCC.}(c), the input-channel window of filter 3 will include the $C_{in}3$ and $C_{in}0$. The benefits of our SCC design compared with GPW can be demonstrated by comparing line 3 and line 2 in Table~\ref{table: comparison with PW and GW.}, where SCC wins in terms of accuracy by better utilizing cross-channel information.

Moreover, another highlight of our SCC design is its flexibility, and there are two parameters are introduced: channel group number $cg$ and input-channel overlap ratio $co$. For instance, we use SCC-cgX-coY\% to denote each filter in the convolution kernel takes $\frac{1}{X}$ number of input channels, while adjacent filters in SCC have y\% overlap in their consumed input channels. 
As exemplified in Figure~\ref{fig: Comparison of PW, GC and SCC.}(c) with the setting of SCC-cg2-co50\%, the input and output channels of the feature map are divided in to two groups, each adjacent filter is overlapped with 50\% input channels. 

\subsection{Implementation Challenges}
While the SCC allows users to explore more algorithmic benefits from DSCs through combining group convolution and overlapping channels, the implementation of this new kernel is challenging in several aspects. 
First,  
existing off-the-shelf high-level CNN building blocks (\textit{e.g.}, cuDNN~\cite{chetlur2014cudnn}, Conv2D in Pytorch) crafted for standard convolution could not be easily adapted for the implementation of SCC due to the quite different ways of conducting convolution operations. Because each filter of SCC relies on different input channels to generate feature map on one output channel, as indicated in Figure~\ref{fig: Comparison of PW, GC and SCC.}. Thus, extra computations are required to pinpoint the corresponding input-channel window at the input feature map before the convolution.   

Second, the implementation of SCC could not well benefit by leveraging the highly-optimized low-level NN primitives, such as General Matrix-Matrix operations (GEMM). The major reason is that SCC requires more fine-grained GEMM operations on the matrix that are essentially skewed in its shape. Because each kernel filter does not cover the identical ``window'' of input channel as its adjacent filter, thus, lacking the reuse of input-channel window. For example, considering a setting with feature map ($56\times56$) in each input channel, the number of input channel is 64 ($C_{in}=64$) and the number of output channel is 128 ($C_{out}=128$). When $cg=2$ (\textit{i.e.}, both input and output channel are divided in to 2 groups), SCC requires 128 times fine-grained GEMM operations between the matrix with shape ($(56\times56)\times32$) and matrix with shape ($32\times1$). Even though stacking these small matrices in to a larger matrix is an alternative solution, the computation and memory costs are also non-trivial. In contrast, existing convolution kernels, such as GPW, can be well benefited from cuBLAS~\cite{cublas} for implementation. Assuming the same setting ($cg=2$) as the above example, GC only requires two GEMM operations between a matrix with shape ($(56\times56)\times32$) and a matrix with shape ($32\times64$), which is clearly efficient compared with that in SCC.

To tackle these challenges, we first propose two implementations by compositing existing Pytorch operators. Further, we orchestrate our highly-efficient implementation with a set of kernel-level designs and optimizations tailored for SCC computation.
We discuss those details in Section~\ref{sect: implementation}.
\section{Implementation} \label{sect: implementation}
\subsection{Pytorch Operator Composition} \label{sect: Pytorch Operator Composition}
\noindent \textbf{Channel-Stack Implementation}
As the most straight-forward solution, we implement SCC by compositing the standard Pytorch operators, such as tensor slicing, concatenation, and standard group convolution. Specifically, there are four major steps, as shown in Figure~\ref{fig: Implementation via Pytorch-operator composition.}(a). The first step (\circled{1}) is to identify the input channels of kernel filters (\textit{i.e.}, the calculation of the index range of each kernel filter, including its starting and ending location). The second step (\circled{2}) is to extract the input feature maps based on the calculated input channel windows from the previous step. Then the third step (\circled{3}) concatenates them into a large feature map long their channel dimension. The fourth step (\circled{4}) is to apply the standard group convolution (such as \texttt{conv2D} in Pytorch) with the number of groups specified as the number of output channels (kernel filters).

However, we could easily find the drawbacks of such an implementation in the follow aspects. First, massively dividing and concatenating tensors (feature maps) in Pytorch is inefficient. It generally requires random indexing and data duplication on tensors, leading to a large amount of data movement that will hurt the performance. 
Second, Pytorch framework lacks support for concurrently executing the above operations, thus, missing the opportunity for parallelization.
Third, the overlapped input channels of each kernel filter incur excessive data redundancy of repeatedly storing the same feature maps, leading to a prohibitively large size of the concatenated tensor that hurdles its applicability on modern GPUs with limited size of memory.

\vspace{4pt}
\noindent \textbf{Convolution-Stack Implementation} 
The second implementation circumvents the ``huge'' concatenated tensor in the above channel-stack implementation by applying convolution operation before concatenating. One major key insight is that the computation on the large concatenated tensor can be decomposed into the more effective computation on a set of small tensors. As illustrated in Figure~\ref{fig: Implementation via Pytorch-operator composition.}(b), instead of simply combining all the extracted features maps, we can pre-build a set of lightweight convolutions, such as GPW1 - GPW4, each of which will generate the feature map for only one kernel filter. Finally, we concatenate these output feature map together. While this solution can largely overcome the third problem of the above channel-stack implementation, it is still hindered by the excessive inefficient Pytorch operations and lack of computation parallelization. 

\subsection{\Mname~Implementation} \label{sect: DeepXplore}
To handle the aforementioned challenges, our \Mname~introduces a set of new designs at forward and backward pass tailored for SCC computation. Moreover, \Mname~identifies the channel-cyclic pattern in SCC that can be effectively applied towards different implementations.

\vspace{3pt}
\noindent \textbf{Output-Centric Forward Pass}
We propose an output-centric forward pass to efficiently handle the SCC forward propagation. 
Adjacent kernel filters are overlapped in terms of their corresponding input channels. 
Note that the last input channel is logically adjacent to the first input channel to form a channel circle. 
In the forward pass, we assign $N\times C_{out} \times F_{w}\times F_{w}$ GPU threads in total to generate each pixel point in the output feature map, and the workload of each thread is just a simple vector-vector multiplication between the weight and the pixels across different input channels but at the same position of the 2D kernel dimension (\textit{i.e.}, the same ($x$, $y$) in $F_{w}\times F_{w}$ space). 
For example, output feature map on channel $C_{1}$ with the shape of $F_{w} \times F_{w}$ will be generated by the elementwise multiplication between input feature maps ($(F_{w}\times F_{w}) \times2$) and weight parameters ($2\times1$) by leveraging $F_w\times F_w$ threads.  

Our proposed forward pass scheme has two major benefits. First, no data duplication for any part of tensors (feature maps) during the whole computation, since each thread only need to fetch its corresponding pixel point across different input channels from the original input tensor directly. Second, better data locality in the input feature map and weight parameters, since adjacent threads within the same GPU block are scheduled to operate on the same output feature map. 
Third, no inter-thread contention, since the computation of each pixel point is independent from each other and is handled by one thread.

\vspace{3pt}
\noindent \textbf{Input-Centric Backward Pass} 
For training our SCC kernel, an efficient backward phase is required for gradient backward propagation to update the model parameters. 
Compared with the forward pass, backward propagation is generally several times higher in computations due to the gradient backpropagation for both the trainable parameters (e.g., weight, and bias) and input feature maps.
The backpropagation of the standard/group convolution can be easily formalized as the high-performance GEMM operation due to the fact that their kernel filters are either fully overlapped (for filters from the same group) or non-overlapped (for filters from different groups) in terms of their input channels. 
However, in our SCC kernel, filters are partially overlapped, which challenges the implementation of gradient backpropagation due to lacking support of any existing optimized operation. 
\begin{figure} [t] \small
    \centering
    \includegraphics[width=0.5\textwidth]{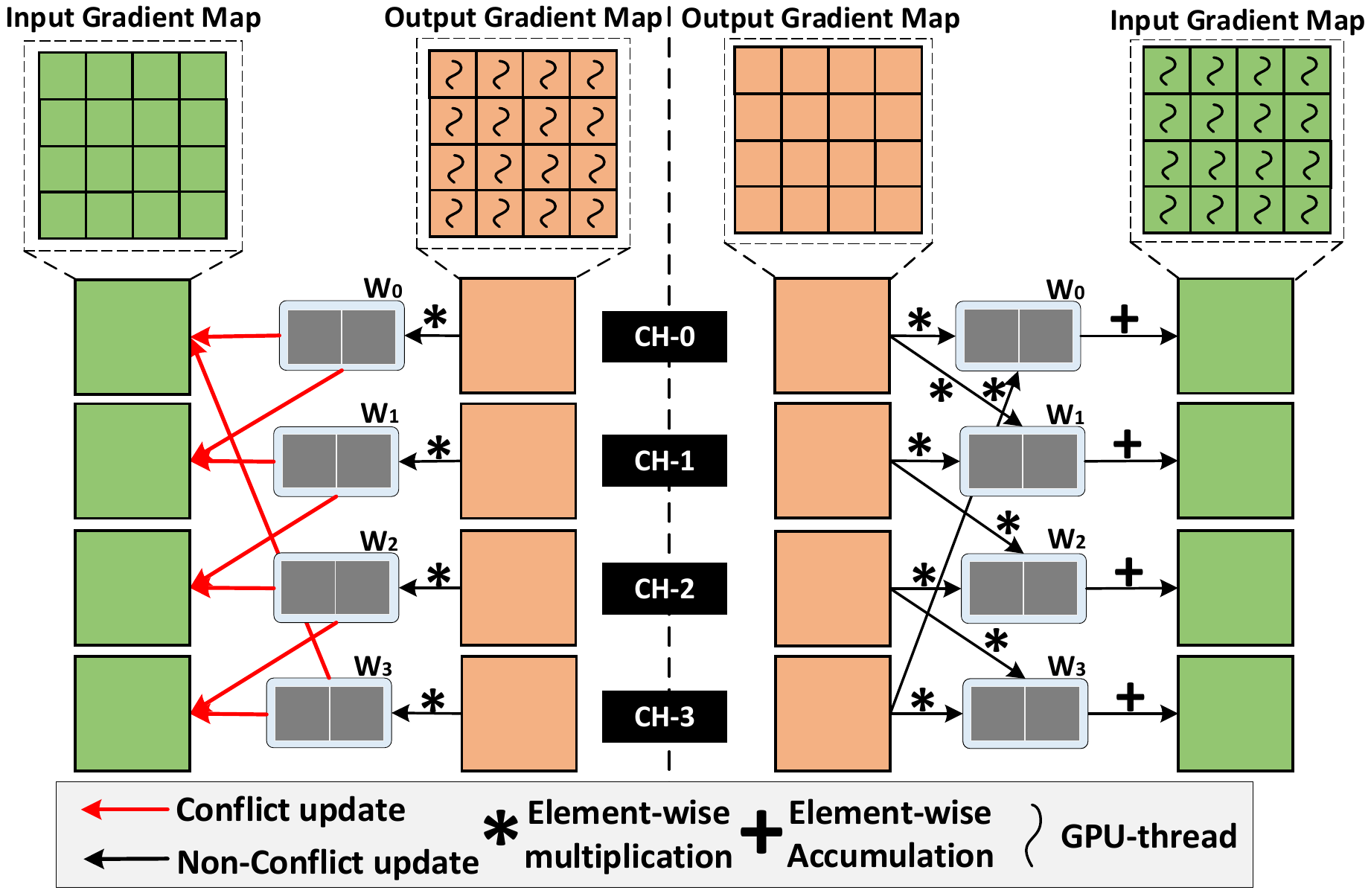}
    \caption{SCC backward computation flow: \textit{output-centric} and \textit{input-centric} design. Note that square boxes stand for points on feature maps, while each curved line inside a box stands for a GPU thread for generating value at that point. ``CH'' is short for ``channel''.}
    \label{fig: Foward and Backward Pass.}
    \vspace{-15pt}
\end{figure}

While there is a straightforward option to simply reverse the forward computation flow of data propagation to the backward flow for gradient propagation. 
As illustrated in Figure~\ref{fig: Foward and Backward Pass.}(a), we assign threads based on the output gradient map as we previous do in the forward pass. 
However, the overlapping of input channel for different filters also leads to the excessive amount of thread-level synchronization during the backward pass. 
The major reason is that the overlapped input-channel feature maps would simultaneously receive the gradients being ``pushed'' from different kernel filters that are operated by different groups of threads. 
Therefore, the computation correctness has to be guaranteed by employing lots of atomic primitives when accumulating gradients for input feature maps. 
For example, the propagation of output gradient map on CH-2 and CH-3 would incur lots of conflicts during the updating of input gradient map on CH-3.

To overcome these issues, we introduce an input-centric backward scheme that assign each thread to generate gradient of each pixel point on input feature map.
As shown in Figure~\ref{fig: Foward and Backward Pass.}(b), we assign threads based on the input gradient map instead of the output gradient map, while the direction of the gradient flow remains unchanged. For example, the input gradient map on CH-3 will ``pull'' gradients from the output gradient maps on CH-2 and CH-3. By harnessing such a input-centric backward scheme, we eliminate the need of using the excessive amount of thread-level synchronization (atomic operations) for the input gradient map update, meanwhile maintaining the correctness of the backward computation.

Note that for both forward and backward computation, we decide not to move forward with GEMM-based (\textit{e.g.}, cuBLAS~\cite{cublas}) solution according to our experimental study.
First, it would lead to launching excessive amount of small GEMM kernels, each of which can not fully utilize the GPU resources; 
Second, cuBLAS~\cite{cublas} no longer supports any function call from the CUDA kernel (decorated with \texttt{\_\_global\_\_}) since CUDA 10.0. Thus, only the host code running on CPU is allowed to invoke \texttt{cublasSgemm}, which largely limits the computation parallelism (even with OpenMP). 

\vspace{3pt}

\noindent \textbf{Channel-Cyclic Optimization}
While SCC brings challenges of implementing forward and backward pass, it also comes with a special pattern -- \textit{channel cyclic}, which can be leveraged for further optimization. The adjacent kernel filters would ``gather'' information from the adjacent ``windows'' of input channels that are overlapped with each other. After several kernel filters, it will encounter the same input-channel window that has been traversed before. For example, as shown in Figure~\ref{fig: Channel-Cyclic.}(a) for the $cg=2$ and $co=50\%$ case, every four kernel filters will share the same set of input-channel window. Similarly, for the $cg=2$ and $co=33\%$ case, as shown in Figure~\ref{fig: Channel-Cyclic.}(b), every three kernel filters will share the same set of input-channel window. To identify such cyclic channel pattern and compute the input-channel indexes of filters within one cycle, we follow the Algorithm~\ref{alg: Identify channel cyclic.}, under the given $C_{in}$, $cg$, and $co$. 

\begin{figure} [t] \small
    \centering
    \includegraphics[width=\columnwidth]{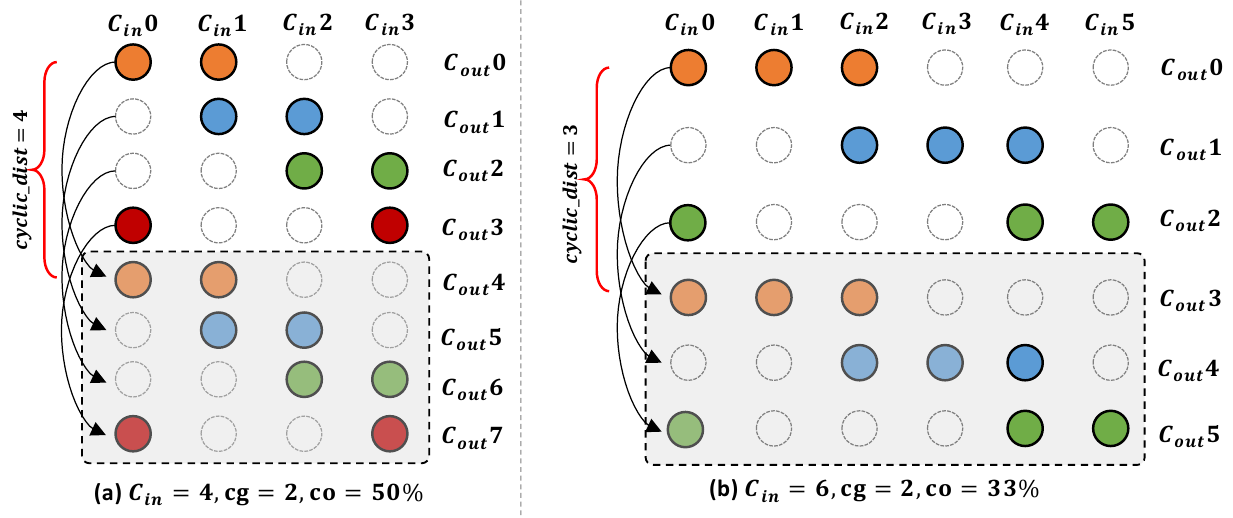}
    \caption{Illustration of the channel-cyclic pattern in SCC.}
    \label{fig: Channel-Cyclic.}
\end{figure}
\begin{algorithm}[h] \small
    \caption{Compute channel indexes of one cycle.}
    \label{alg: Identify channel cyclic.}
    $channel\_map = \{\}$ \;
    $group\_width = input\_channel//num\_groups$\;
    $start, end = 0, group\_width$\;
    $start\_v, end\_v = start, end$\;
    $cyclic\_dist = 0$\;
    \For {$oid=0; \ oid < output\_channel; \ oid++$}{
        $item = (start, end)$\;
        \If {$item$ \textbf{not} in $channel\_map$}{
          $channel\_map.add(item)$\;
          $cyclic\_dist++$\;
        }
        \Else{
            \textbf{break}\;
        }
        $start\_v = end\_v - int(overlap * group\_width)$\;
        $end\_v = start\_v + group\_width$\;
        $start = start\_v\ \% \ input\_channel$\;
        $end = end\_v \ \% \ input\_channel$\;
    }
\end{algorithm}

Cyclic-channel pattern in SCC can be leveraged to optimize our proposed implementations in the above sections.
First, it can be applied to the implementations of Pytorch operator composition to reduce the repetitive tensor manipulations and save the memory. 
\begin{figure} [t] \small
    \centering
    \includegraphics[width=0.5\textwidth]{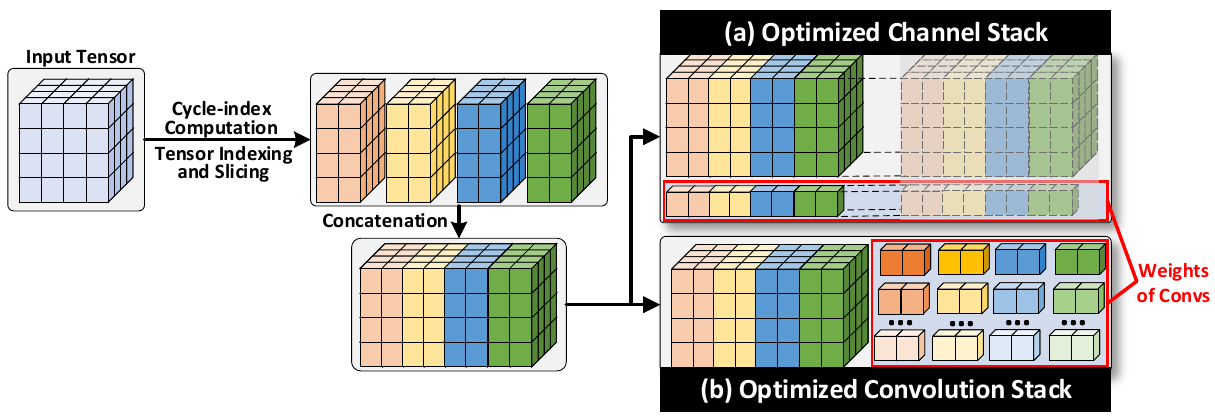}
    \caption{Case study of cyclic-channel optimization on Pytorch-based implementations for $C_{in}=4$, $cg=2$, $co=50\%$.}
    \label{fig: cyclic optimization on Pytorch.}
    \vspace{-15pt}
\end{figure}
Specifically, in the channel-stack implementation, we only need to identify and concatenate the first cycle of input feature maps. 
Then for the settings that are composed of multiple cycles, we just need to duplicate the concatenated feature map of the first cycle several times and merge them all together along the channel dimension, as exemplified in Figure~\ref{fig: cyclic optimization on Pytorch.}(a). 
This can avoid the high-cost tensor indexing and slicing operations on the original feature map that have been carried out before. 
In the convolution-stack implementation, we can largely save the memory by keeping the concatenated tensor of the input feature maps in the first cycle, since all the remaining cycles would maintain the same input information, as illustrated in Figure~\ref{fig: cyclic optimization on Pytorch.}(b). 
And we just need to pile enough GPW convolutions (the number of which equals the output channels) that will be trained to extract different information from the same input. 
\begin{algorithm}[h] \small
    \caption{\Mname~Channel-cyclic Optimization}
    \label{alg: Get the input channel range.}
    \SetAlgoLined
    $thread\_id = blockIdx.x *blockDim.x + threadIdx.x$\;
    $opt\_channel\_id = get\_output\_channel(thread\_id)$\;
    $idx = output\_channel\_id \ \% \ cyclic\_dist$\;
    $start, end = channel\_map[idx]$\;
\end{algorithm}

Second, it can be leveraged to optimize our \Mname~by reusing the indexes of the input channel during our SCC-tailored forward pass. 
Therefore, the recurrent index range only need to be calculated once and will be stored into a map/dictionary like object for the following filters,  
as illustrated in Algorithm~\ref{alg: Get the input channel range.}. 
$Line \ 1$ gets the corresponding kernel filter index ($opt\_channel\_id$) based on the GPU thread ID. Then based on the $opt\_channel\_id$ and the $cyclic\_dist$ from the Algorithm~\ref{alg: Identify channel cyclic.}, we can get the $idx$ of feature map to fetch the corresponding starting and ending position of input channels for the filter.

\section{Evaluation} \label{sect: evaluation}
In this section, we conduct a set of intensive experiments on \Mname~in terms of its accuracy and the runtime performance.

\subsection{Experimental Setup}
\underline{\textit{\textbf{Datasets}}}
We use CIFAR-10~\cite{cifar10} and ImageNet~\cite{deng2009imagenet} dataset for evaluation. CIFAR-10 consists of 60,000 32$\times$32 color images in 10 classes, with 6,000 images per class. ImageNet is a large dataset of over 14 million images with up to 1,000 output classes, and it has been widely used for many computer vision research.

\underline{\textit{\textbf{CNN Models}}} We run comprehensive experiments on the state-of-the-art CNN models (VGG16~\cite{vgg} and VGG19~\cite{vgg}, MobileNet~\cite{mobilenet_2017_howard}, ResNet18~\cite{he2016deep} and ResNet50~\cite{he2016deep}. The major reasons of choosing these CNN models are 1) VGG16 and VGG19 are two most classic CNNs with linearly stacked layers; 2) MobileNet is the typical lightweight model with depth-separable convolution (DW+PW) blocks; 3) ResNet18 and ResNet50 is the representative model with the non-linearly stacked layers (residual connections).

\underline{\textit{\textbf{Implementations}}} We include three implementations for comparison: 
1) \textbf{Pytorch-Base}: the baseline Pytorch implementation using channel-stack design (\textbf{CHS}) without channel-cyclic (\textbf{CC}) optimization; 
2) \textbf{Pytorch-Opt}: the optimized Pytorch implementation using convolution-stack (\textbf{COS}) design with CC optimization; 
3) \textbf{\Mname}: \Mname~implementation comes with our efficient forward/backward pass design and CC optimization. 
Note that we omit the Pytorch implementation with CHS design and CC optimization, since CHS design still requires replicating the identified recurrent feature maps to build a large tensor before passing through the group convolution, which is identical to the Pytorch-Base implementation in terms of computation and memory complexity. 

\underline{\textit{\textbf{Platforms \& Tools}}} Our major evaluation platform is a server with an Intel Xeon Platinum 8168 (2.7 GHz, 24 cores), 188GB host memory and a Tesla V100 GPU (5,120 CUDA cores, Memory: 32GB, Peak Single-Precision: 15.7 TFLOPs).
Note that to measure the end-to-end CNNs training performance, we use the Python \textit{time} library and calculate the averaged running time of 100 measurements under the same setting.
For kernel detailed metric analysis, such as GPU memory consumption, we use NVProf profiling tool from Nvidia. 


\subsection{Algorithmic Performance} 
\label{sect: Algorithmic Performance}
In this experiment, we will first show the overall accuracy performance of different models optimized by \Mname~on CIFAR-10 in terms of computation (FLOPs), parameter saving and accuracy. 
We then apply our SCC kernel on ResNet50 and evaluate it on ImageNet dataset to show its applicability towards more complicated model on the large dataset.
Moreover, we demonstrate the benefits of our SCC kernel design by using MobileNet on CIFAR-10 for a detailed study with the different number of groups ($cg$) and the channel overlapping ratios ($co$). Note that the value of $cg$ should respect 1) the smallest channel number ($64$ for our selected models) of the convolution layers (excluding the input layer, which is usually 3 for RGB image), 2) balancing the parameter/computation reduction and the model accuracy. Our empirical study shows that $cg > 8$ would lead to significant accuracy degradation.
\begin{table}[h]
\centering
\caption{Accuracy comparison of CNNs on CIFAR-10.}
\scalebox{0.8}{
\begin{tabular}{c c r r c}
\toprule
\multicolumn{1}{l}{\textbf{Model}} &
  \textbf{Implementation} &
  \multicolumn{1}{l}{\textbf{MFPLOS}} &
  \multicolumn{1}{l}{\textbf{Param. (M)}} &
  \multicolumn{1}{l}{\textbf{Acc. (\%)}} \\ \hline \hline
                            & Origin   & 314.16  & 14.73M                     & 92.64 \\
\multirow{-2}{*}{VGG16}     & DSXplore & 21.85   & 0.87M                      & 92.60 \\ \hline
                            & Origin   & 399.17  & 20.04M                     & 93.88 \\ 
\multirow{-2}{*}{VGG19}     & DSXplore & 26.92   & 1.19M                      & 92.71 \\ \hline
                            & Origin   & 50.00   & 6.17M                      & 92.05 \\ 
\multirow{-2}{*}{MobileNet} & DSXplore & 30.00   & 0.59M                      & 92.56 \\ \hline
                            & Origin   & 255.89  & 11.17M                     & 95.75 \\ 
\multirow{-2}{*}{ResNet18}  & DSXplore & 43.99   & 0.84M                      & 94.44 \\ \hline
                            & Origin   & 1297.80 & 23.52M                     & 95.82 \\ 
\multirow{-2}{*}{ResNet50}  & DSXplore & 735.79  & 12.87M                     & 95.12 \\
\bottomrule
\end{tabular}}
\label{table: Accuracy Comparison of CNN models on CIFAR-10.}
\end{table}

\begin{table}[h] \small
    \caption{Accuracy comparison (\textbf{ImageNet}) for ResNet50.}
    \label{tab: Performance comparison ImageNet of ResNet50 Variants.}
    \centering
    \scalebox{0.9}{
    \begin{tabular}{cccc}
        \toprule
        Network & MFLOPs & Param. & Acc.(\%)\\
        \midrule
        \midrule 
        Origin        & 4130              & 23.67M          & 76.56
        \\\midrule
       \textbf{\Mname}    & \textbf{2550}     & \textbf{14.34M} & \textbf{75.91}
        \\
        \bottomrule
    \end{tabular}}
    \vspace{-13pt}
\end{table}
\begin{table}[h]
    \caption{Comparison of different settings on MobileNet.}
    \label{tab: MobileNet}
    \centering
    \scalebox{0.83}{
    \begin{tabular}{cccc}
        \toprule
        Network & MFLOPs & Param. & Acc.(\%)\\
        \midrule
        \midrule 
        Baseline (DW+PW)        & 50     & 6.17M  & 92.05          \\ \midrule
        \textbf{DW+GPW-cg2}        & \textbf{30}     & \textbf{0.59M}  & \textbf{90.11}          \\\midrule
        \textbf{DW+GPW-cg4}        & \textbf{20}     & \textbf{0.32M} & \textbf{88.88}          \\\midrule
        \textbf{DW+GPW-cg8}        & \textbf{10}     & \textbf{0.18M} & \textbf{82.69}          \\\midrule
        \textbf{DW+SCC-cg2-co33\%}  & \textbf{30}     & \textbf{0.59M}  & \textbf{91.20}          \\\midrule
        \textbf{DW+SCC-cg2-co50\%}  & \textbf{30}     & \textbf{0.59M}  & \textbf{92.56}          \\\midrule
        \textbf{DW+SCC-cg4-co33\%}  & \textbf{20}     & \textbf{0.32M} & \textbf{91.71}          \\\midrule
        \textbf{DW+SCC-cg4-co50\%}  & \textbf{20}     & \textbf{0.32M} & \textbf{91.39}          \\\midrule
        \textbf{DW+SCC-cg8-co33\%}  & \textbf{10}     & \textbf{0.18M} & \textbf{90.71}          \\\midrule
        \textbf{DW+SCC-cg8-co50\%}  & \textbf{10}     & \textbf{0.18M} & \textbf{90.25} \\
        \bottomrule
    \end{tabular}}
    \vspace{-5pt}
\end{table}
\begin{figure*}[t] \small
    \centering
    \subfloat{\includegraphics[width=0.5\textwidth]{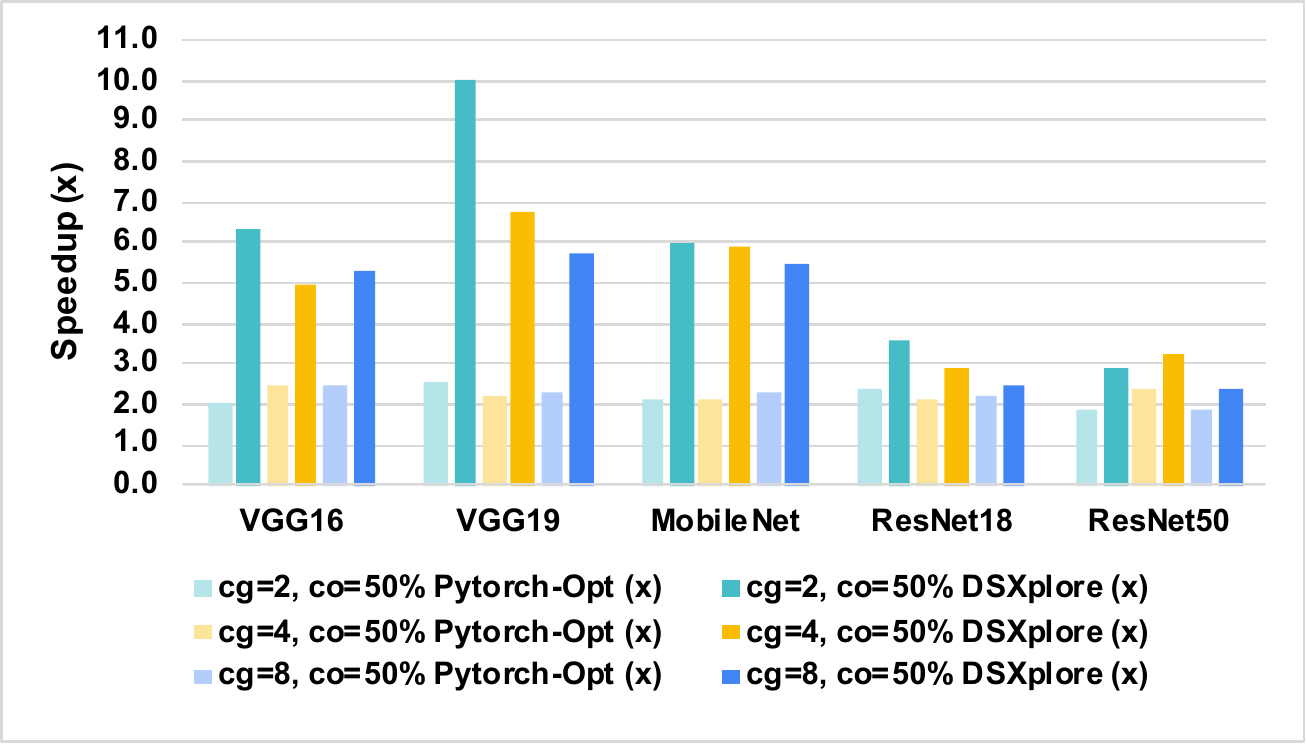}}
    \subfloat{\includegraphics[width=0.5\textwidth]{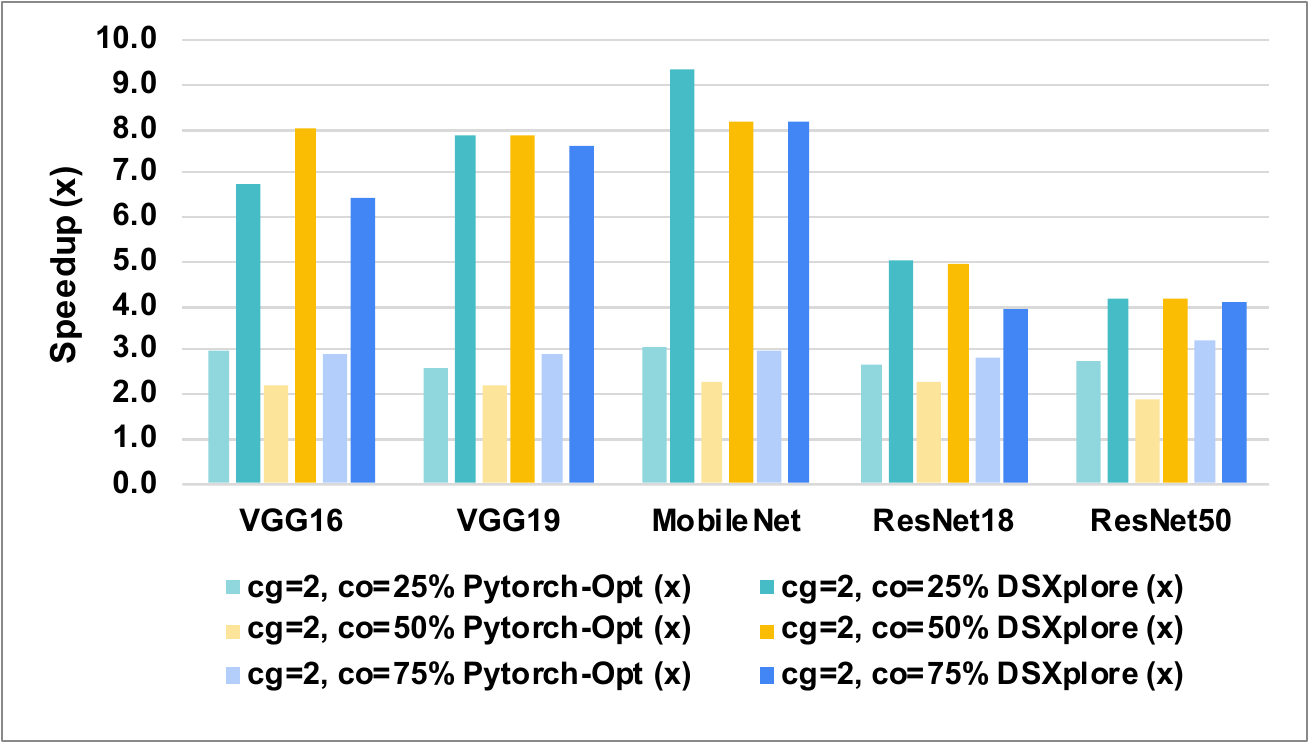}}
    \caption{Runtime performance comparison on CIFAR10. Note that speedup is normalized \textit{w.r.t.} \textbf{Pytorch-Base} Implementation.}
    \vspace{-5pt}
    \label{fig: Runtime Performance Comparison on CIFAR10}
\end{figure*}
\begin{figure*}[t] \small
    \centering
    \subfloat{\includegraphics[width=0.5\textwidth]{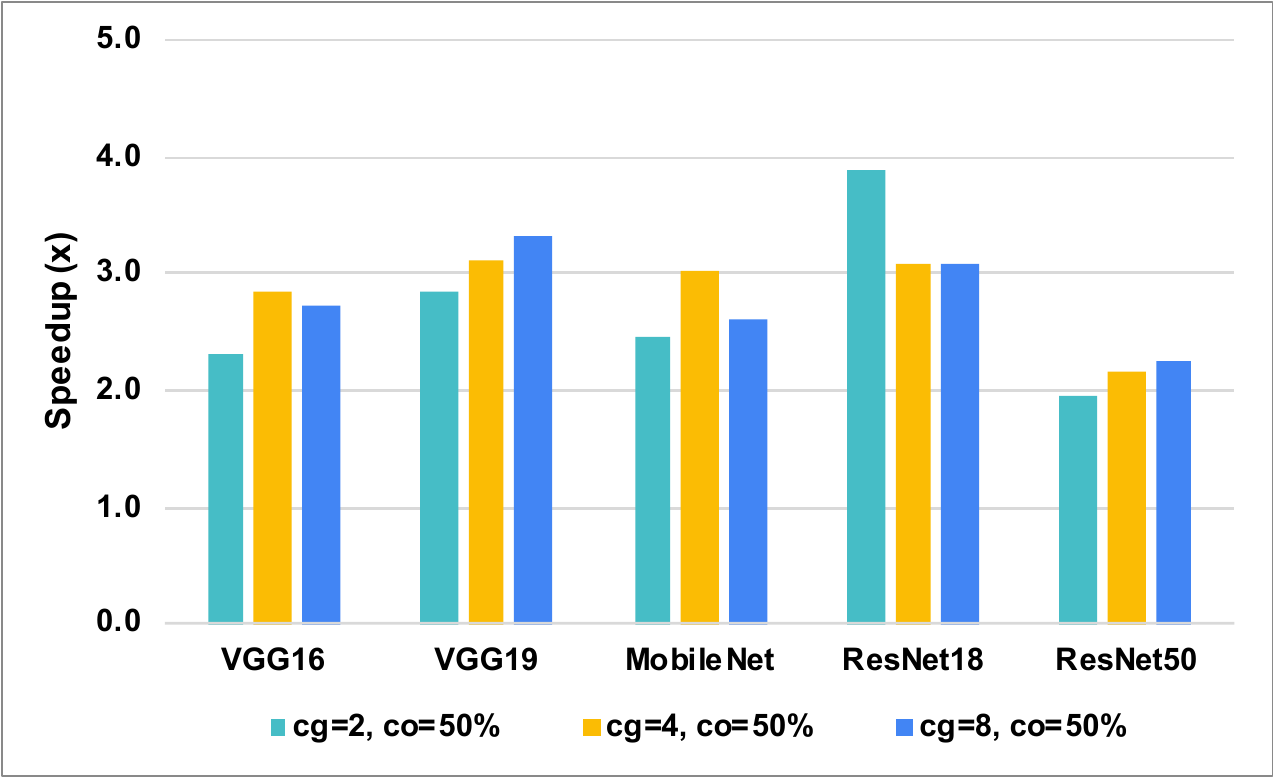}}
    \subfloat{\includegraphics[width=0.5\textwidth]{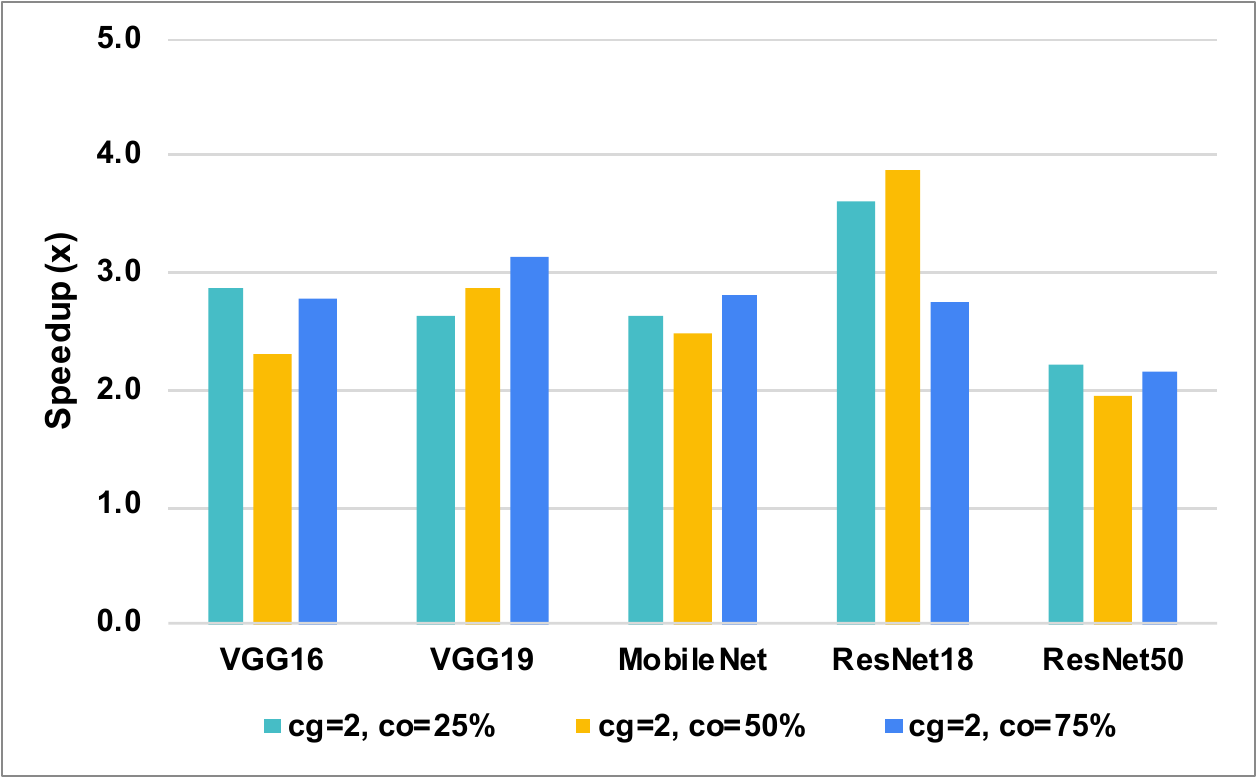}}
    \caption{Runtime performance comparison on ImageNet. Note that speedup is normalized \textit{w.r.t.} \textbf{Pytorch-Opt} Implementation.}
    \vspace{-5pt}
    \label{fig: Runtime Performance Comparison on ImageNet}
\end{figure*}

\vspace{3pt}
\noindent \textbf{Overall Accuracy}
As shown in Table~\ref{table: Accuracy Comparison of CNN models on CIFAR-10.}, DSXplore optimized CNNs can strike a good balance between the model accuracy and the computation/parameter size. 
Overall, \Mname~can save 70.48\% FLOPs in computation and 83.27\% in parameter size on average, meanwhile maintaining accuracy to a great extent. 
This is because that \Mname~leverages the SCC design that can maximize the value from the input information across different channels. 
We also notice that on VGG16 and VGG19, \Mname~can save more than 90\% FLOPs and more than 90\% parameters. 
The major reason is that the original VGG models rely on standard convolutions that would carry lots of redundant computations and parameters that contribute minor to the final prediction. 
In contrast, our SCC design in \Mname~can effectively capture the key factors (spatial and cross-channel information) that contributes most to the model prediction capability, thus, reducing the computation and parameters without much compromising accuracy.
Besides, for the complicated ResNet50 model on the challenging ImageNet dataset (Table~\ref{tab: Performance comparison ImageNet of ResNet50 Variants.}), our SCC design can still reduce the FLOPs and parameters with up to 38.25\% and 39.41\% compared with the original model while maintaining accuracy.

\vspace{3pt}
\noindent \textbf{Detailed Analysis}
A detailed studies on MobileNet (Table~\ref{tab: MobileNet}) further demonstrates the advantage of our SCC design in \Mname~in terms of a better trade-off between the model efficiency and the prediction accuracy. We try three different group numbers $cg\in\{2, 4, 8\}$), as well as two overlapping ratios $co\in \{33\%, 50\%\}$. 
Our model with DW+SCC-cg2-co50\% achieves a better accuracy compared to the baseline DW+PW model while saving about 40.00\% FLOPs and 90.43\% parameters. 
The reason is that under the setting of the small group number (\textit{e.g.}, 2), our SCC-based channel sliding help filters learn different information by watching different channels, while the DW+PW cases consistently looks at the same input channels. 
Note that in the larger group number (\textit{e.g.}, 8), such benefits would be largely offset by the significantly reduced parameters.
With an increase in the group number, we observe a significant reduction in both computational cost and parameter usage, along with a slight degradation in prediction accuracy. 
This aligns well with our expectation that the channel-group number $cg$ determines the number of input channels that GPW or SCC would take, and thus also decides the number of computations and parameters of the model. 
Besides, SCC kernel design consistently outperforms the ones without overlapping under the same number of groups ($cg$). For example, our new design (DW+SCC-cg4-co33\%) outperforms DW+GPW-cg4 with 2.83\% better accuracy. 

\subsection{Runtime Performance}
In this section, we compare \Mname~with Pytorch-Base and Pytorch-Opt on CIFAR-10 and ImageNet for training across different CNNs, including VGG16, VGG19, MobileNet, ResNet18, and ResNet50. We use two set of settings for better coverage, where the first type of settings is $cg\in\{2,4,8\}$ and $co=50\%$ while the second type of settings is $cg=2$ and $co\in\{25\%,50\%,75\%\}$.

As shown in Figure~\ref{fig: Runtime Performance Comparison on CIFAR10}, across all two types of settings, \Mname~consistently outperforms Pytorch-Base and Pytorch-Opt with average $5.68\times$ and $2.34\times$ speedup, respectively. 
We also notice that Pytorch-Opt is faster than Pytorch-Base, since our effective convolution-stack design and channel-cyclic optimization can be leveraged to reduce the memory overhead and excessive data movement (tensor slicing and concatenation), such an optimization can deliver $1.86\times$ to $3.20\times$ speedup compared with the Pytorch-Base implementation. \Mname~further boosts the performance on top of that by introducing channel-wise parallelism, which delivers an additional $1.11\times$ to $3.97\times$ speedup on average compared with Pytorch-Opt. 
Another key observation is that on VGG16 and VGG19, the performance benefits is more significant compared with ResNet18 and ResNet50.
The major reason is that VGG models mainly rely on standard convolutions (carrying high computation complexity and large number of parameters) as the major building blocks, while the ResNet models would use either ``Basic Blocks'' or ``Bottleneck Blocks'' as the major building blocks. 
These ``Blocks'' include additional convolutions that are already lightweight (such as the dual PW convolutions in Bottleneck Block) and no need to be replaced, thus, replacing the standard convolution of these blocks would have relatively lower impact.
Moreover, under the same overlapping ratio, the increase of the $cg$ would give more advantage to the Pytorch-based implementations, since the computation of each group convolution is reduced due to the smaller number of channels in each group. 
On the other hand, however, as notated in Section~\ref{sect: Algorithmic Performance}, the larger $cg$ would lower the model accuracy. Therefore, we prefer lower value of $cg$ in most settings.

\vspace{-2pt}
In the comparison on ImageNet for the same set of CNNs, Pytorch-Base cannot even to run due to the excessive amount of the memory consumption. Therefore, we choose the Pytorch-Opt as the baseline for speedup normalization. As shown in Figure~\ref{fig: Runtime Performance Comparison on ImageNet}, \Mname~outperforms Pytorch-Opt with $1.95\times$ to $3.88\times$ speedup. This also demonstrate the scalability of \Mname~by effectively exploring the computation parallelism. 

\begin{figure} [t] \small
    \centering
    \includegraphics[width=0.8\columnwidth]{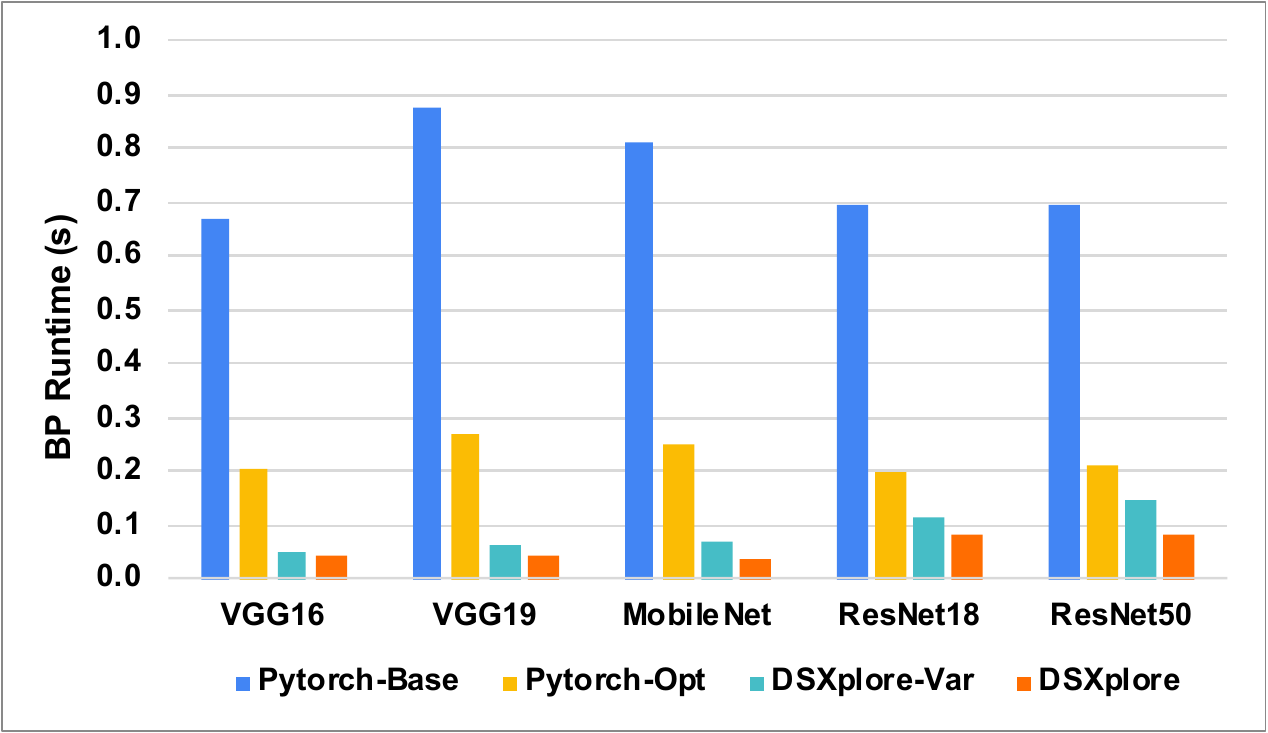}
    \caption{Back-propagation optimization.}
    \vspace{-15pt}
    \label{fig: input-centric Back-Propagation Optimization.}
\end{figure}

\begin{figure} [t] \small
    \centering
    \includegraphics[width=0.8\columnwidth]{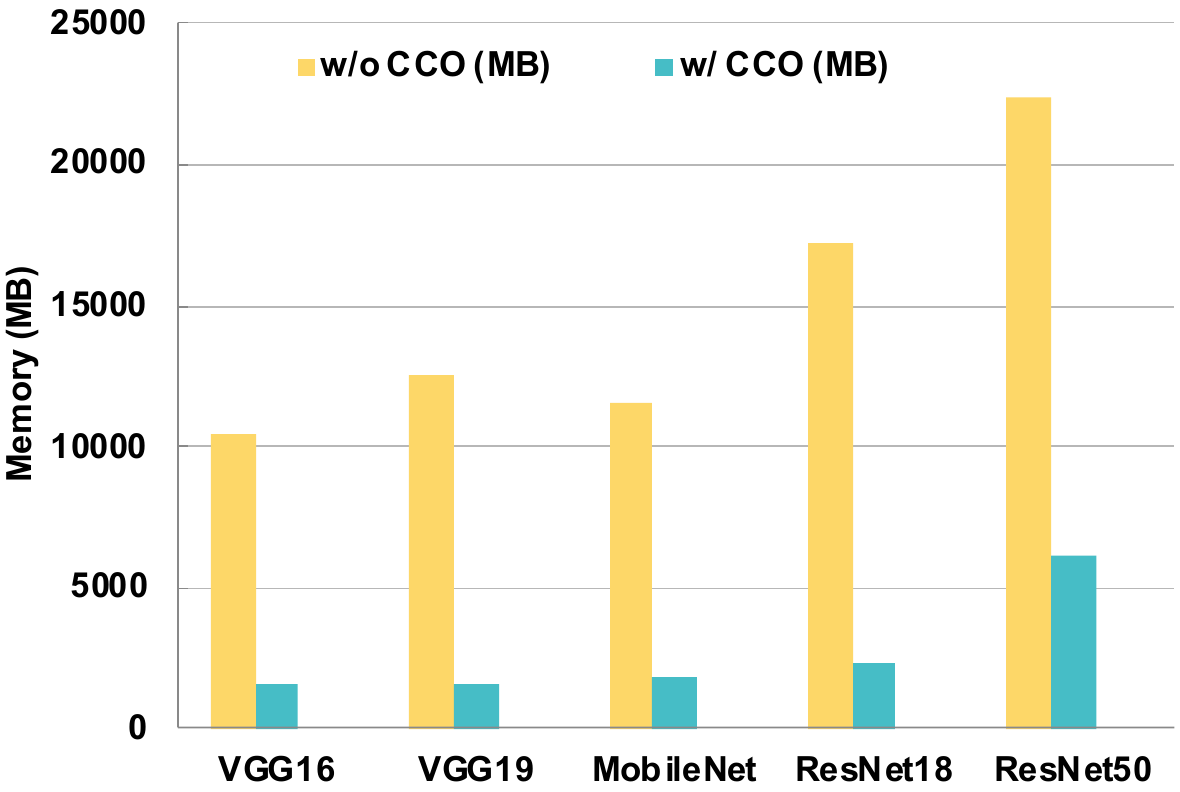}
    \caption{Channel-cyclic optimization.}
    \vspace{-10pt}
    \label{fig: Channel-CyclicOptimization.}
\end{figure}

\subsection{Additional Studies}
\vspace{2pt}
\textbf{Input-centric Backward Design}
To demonstrate the benefits of our input-centric backward optimization, we leverage four implementations (\textbf{Pytorch-Base}, \textbf{Pytorch-Opt}, a \Mname~variant (\textbf{\Mname-Var}) with output-centric backward, which simply reverses the forward propagation flow for backward gradient propagation), and \textbf{\Mname}~(with input-centric backward). 
We evaluate the time of backward gradient propagation only.
As shown in Figure~\ref{fig: input-centric Back-Propagation Optimization.}, our input-centric backward achieves average $15.03\times$, $4.55\times$, $1.55\times$ speedup compared with Pytorch-Base, Pytorch-Opt and \Mname-Var, respectively. The advantage of \Mname~are two-folds: first, compared with Pytorch implementations, \Mname~with output-centric backward propagation (\Mname-Var) and input-centric backward propagation (\Mname) can explore the computation parallelism, meanwhile reducing lots of unnecessary data manipulation; 
second, compared with the \Mname-Var, \Mname~can significantly reduce the atomic operations (more than 90\% on average) on updating the gradient of the input tensor  based on our kernel profiling via NVProf. 


\textbf{Cyclic-Channel Optimization}
As another key technical contribution, our CC optimization can effectively reduce the memory overhead from 72.88\% to 83.33\%, as illustrated in Figure~\ref{fig: Channel-CyclicOptimization.}. This is because the pattern of repeated channel will occur periodically, as described in Section~\ref{sect: DeepXplore}. And we only need to store them once instead of extracting and concatenating them all. To this end, we can avoid most of the data manipulation (\textit{e.g.}, tensor slicing) and movement (\textit{e.g.}, tensor concatenation). What also worth noticing is that the impact of such optimization would also depend on the value of $co$ we choose. Therefore, the cyclic distance would be largely different. For example, when $co=50\%$ with $C_{in}=4$, we will have the cyclic distance of 4. When $co=25\%$ with $C_{in}=6$, we will have the cyclic distance as 3. This would also determine the size of concatenated feature map to be stored, thus, affecting the overall memory consumption. 
 \begin{figure} [t] \small
    \centering
    \includegraphics[width=0.8\columnwidth]{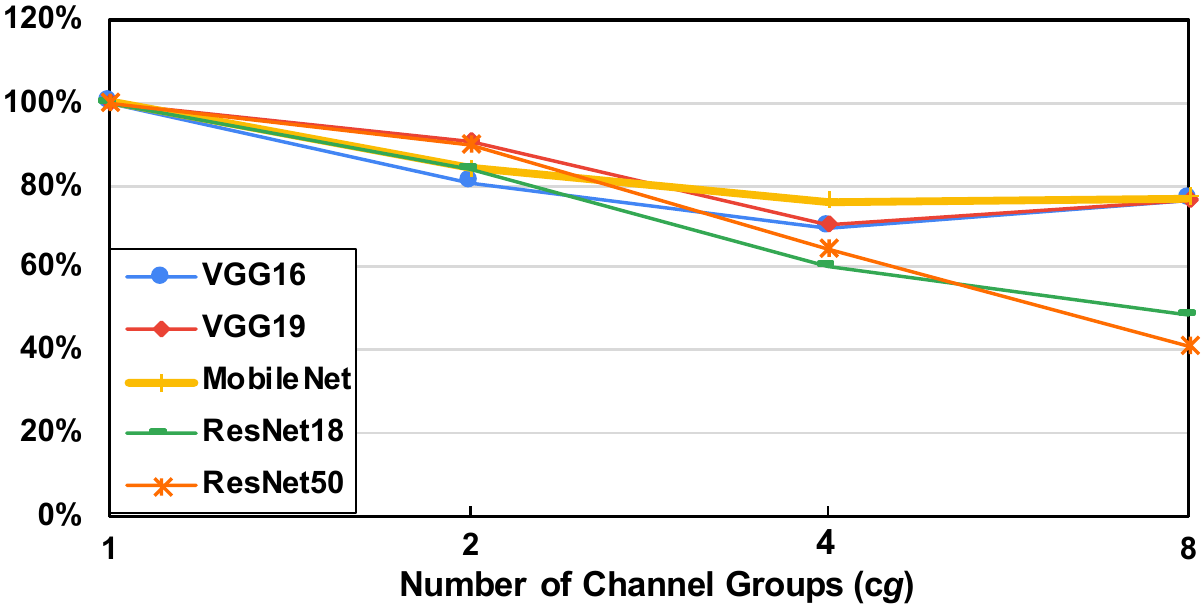}
    \caption{The performance impact of the number of groups ($cg$). \small{Note that we set $co=50\%$ and the runtime is normalized \textit{w.r.t} the performance at $cg=1$}.}
    \vspace{-15pt}
    \label{fig: Number of Groups}
\end{figure}

\vspace{3pt} 
\textbf{Number of Groups} 
As one of the features that is shared with the existing group convolution, our SCC divides the input and output channels into different groups based on the user-provided group numbers. 
Experiments from the detailed analysis of previous Section~\ref{sect: Algorithmic Performance} already gives us the idea about its impact on model accuracy, and this study will help us to analyze its influence on the overall runtime performance in the end-to-end training.
As shown in Figure~\ref{fig: Number of Groups}, the increase of the number of groups will lead to the reduction of the runtime, since with more channel groups, the corresponding group size (the number of input channels) required for each output channel will decrease. 
Therefore, the overall running time will decrease. 
Meanwhile, as discussed in the above section, the more number of groups will also leads to a slightly decrease of accuracy and reduction in parameter/computation costs. 
Therefore, in practice, we should balance the runtime performance and the model accuracy performance when choosing the value of $cg$.
 \begin{figure} [t] \small
    \centering
    \includegraphics[width=0.8\columnwidth]{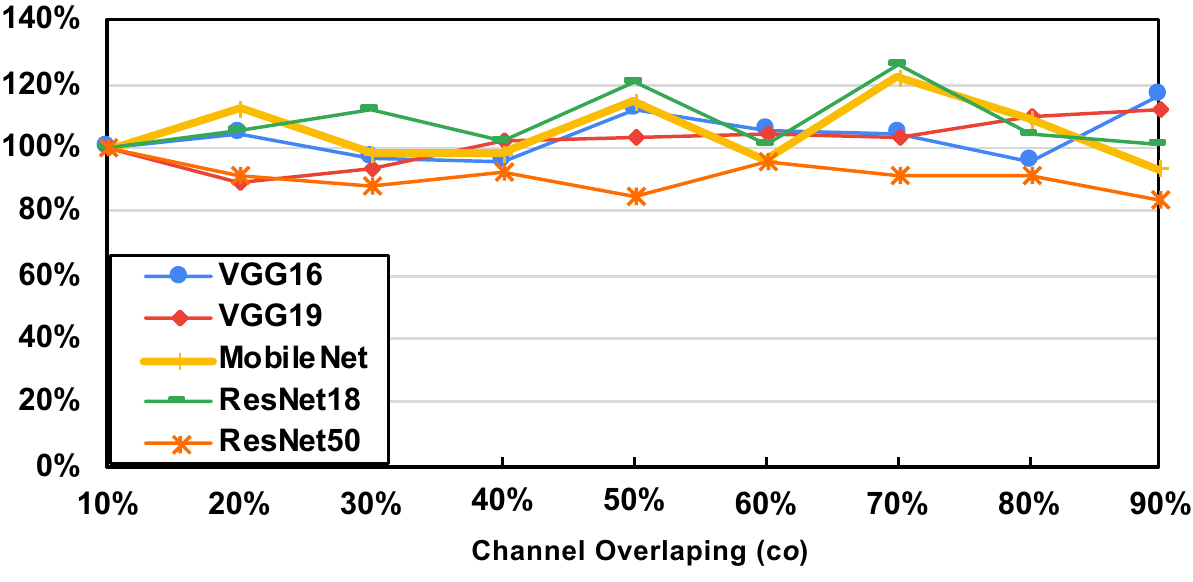}
    \vspace{-5pt}
    \caption{The performance impact of the input-channel overlapping ratio ($co$). \small{Note that we set $cg=2$ and the runtime is normalized \textit{w.r.t} the performance at $co=10\%$}.} 
    \vspace{-10pt}
    \label{fig: The Channel Overlapping Ratio}
\end{figure}

\vspace{3pt} \textbf{Input-Channel Overlapping}
As the key feature that distinguishes SCC from the existing factorized convolution kernels, the overlapping ($co$) of the filters' input channel facilitate the information fusion across different channels. 
Experiments from the detailed analysis of previous Section~\ref{sect: Algorithmic Performance} already gives us the idea about its impact on model accuracy, while this experimental study, on the other side, aims to help us understand how different choices of $co$ would impact our SCC kernel performance. 
As shown in Figure~\ref{fig: The Channel Overlapping Ratio}, the change of $co$ for the adjacent sliding channel units does not show evident impact on the runtime, since the overlapping ratio will not change the workloads assigned to different threads during the forward and backward pass. Even though there are some fluctuation of the running time, and it is mostly caused by the data reuse when choosing different $co$, which is also a very minor impact compared with the change of $cg$. 

\vspace{3pt} \textbf{Training Batch Size}
As one of the most important factor of training CNNs, the batch size would impose profound impact on the training, including the convergence rate, model accuracy, and training speed. 
In general, the larger batch size would lead to shorter training time. However, it may also degrade the model accuracy performance. 
In this experiment, we consider different batch sizes ranging from 16 to 1024 with the increase step by the power of 2. 
We select three CNNs: VGG16, MobileNet, and ResNet18, which represents different types of CNN architectures for studies under the most common settings of $cg=2$ and $co=50\%$. 
As shown in Figure~\ref{fig: Training batch size on performance.}, the overall trend of the running time will increase with respect to the increase of the batch size. We also observe that within a certain range of batch size (less than 128), the increase of batch size does not lead to the evident increase of the running time. 
This is because of not enough active threads to saturate the available GPU Streaming Processors (SMs) to support the fully parallelized forward and backward computation. However, when the batch size becomes even larger, the active threads will increase correspondingly, which are more likely to compete with each other at given the number of SMs for execution. 
 \begin{figure} [t] \small
    \centering
    \includegraphics[width=0.8\columnwidth]{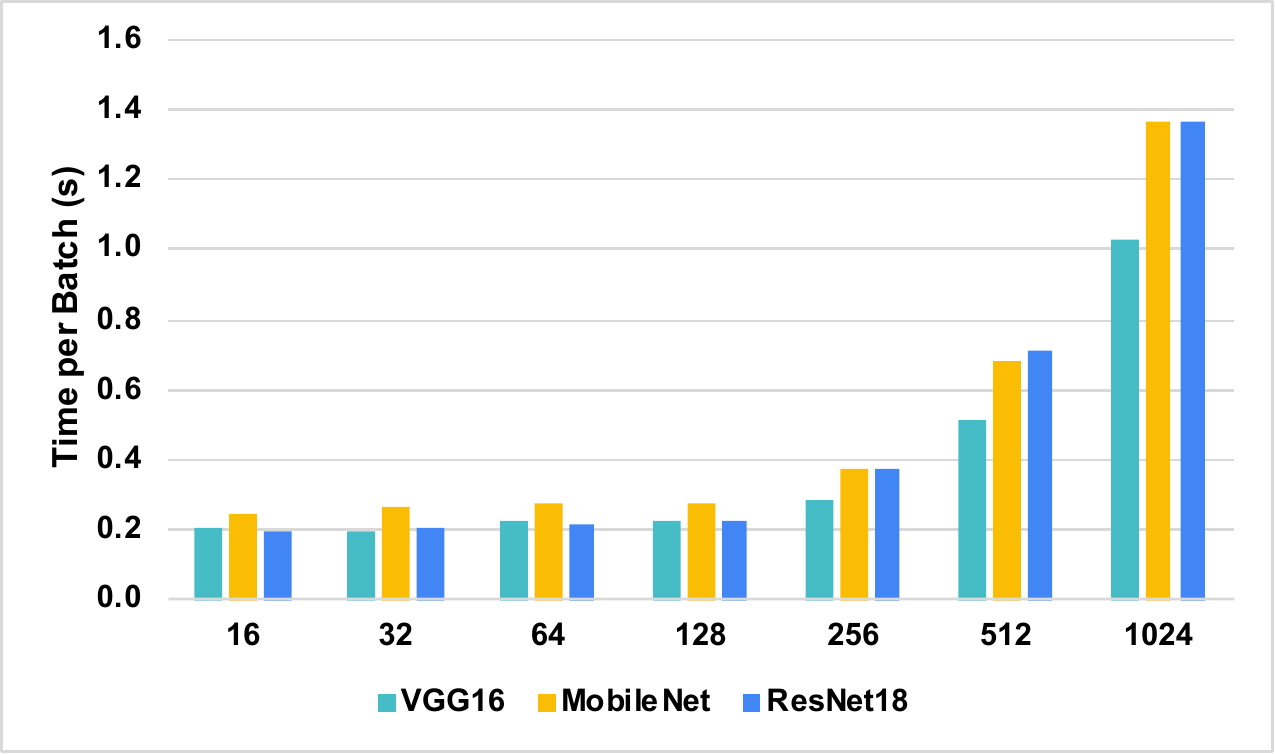} 
    \caption{Impact of batch size on training performance.} 
    \vspace{-5pt}
    \label{fig: Training batch size on performance.}
\end{figure}
 
\vspace{2pt} 
\textbf{Inference Performance}
We compare the inference performance between \Mname~with cg2-o50\%~and DW+GPW-cg2 on VGG16 for CIFAR-10. 
Note that different from CNN training that generally prefers large batch size, CNN inference is commonly working on settings with smaller batch size and focuses on absolute latency. 
Therefore, we choose the common batch size for CNN inference ranging from 16 to 256. 
Note that we report the end-to-end averaged runtime of each batch by processing 100 batches.
As shown in Table~\ref{tbl: Inference Performance}, \Mname~can achieve comparable performance with the DW+GPW, which is built with existing highly engineered libraries, such as cuDNN~\cite{chetlur2014cudnn}. 
This also demonstrates the practicability of \Mname~in real-world deep-learning applications.
\begin{table}[h]
\centering
\caption{Inference performance comparison with DW+GPW.}
\vspace{-3pt}
\scalebox{0.8}{
\begin{tabular}{|r|r|r|}
\hline
\textbf{Batch Size} & \textbf{DW+GPW (ms)} & \textbf{DSXplore (ms)}
\\ 
\hline
\hline
\textbf{16}  & 6      & 8  \\ \hline
\textbf{32}  & 10     & 11 \\ \hline
\textbf{64}  & 10     & 16 \\ \hline
\textbf{128} & 17     & 28 \\ \hline
\textbf{256} & 79     & 75 \\ \hline
\textbf{512} & 90     & 79 \\ \hline
\end{tabular}}
\label{tbl: Inference Performance}
\vspace{-5pt}
\end{table}

\vspace{3pt}
\textbf{Multi-GPU Scalability} 
As another major advancement of the modern CNN training, multi-GPU support is winning lots of attentions from both the industry and research field. In this experiment, we further demonstrate the applicability of \Mname~under the multi-GPU setting. Specifically, we use the same three CNN model as the last experiments for study. 
We start from the single-device (1-GPU) setting, and choose the performance of the 1-GPU as the baseline for performance normalization and comparison for multi-GPU ({2,3,4}-GPU) settings. 
As shown in Figure~\ref{fig: Multi-GPU Scalability.}, the overall trend of the speedup performance is increasing with the increase of the GPUs. 
And we also observe that the speedup performance is also approximate to the linear speedup when the number of GPUs becomes even larger, for example, the 4-GPU setting, whereas fewer GPUs (such as 2, 3) the benefits of multi-GPUs would likely to be offset by the inter-device communication for data transferring and gradient synchronization.
 \begin{figure} [h] \small
    \centering
    \includegraphics[width=0.8\columnwidth]{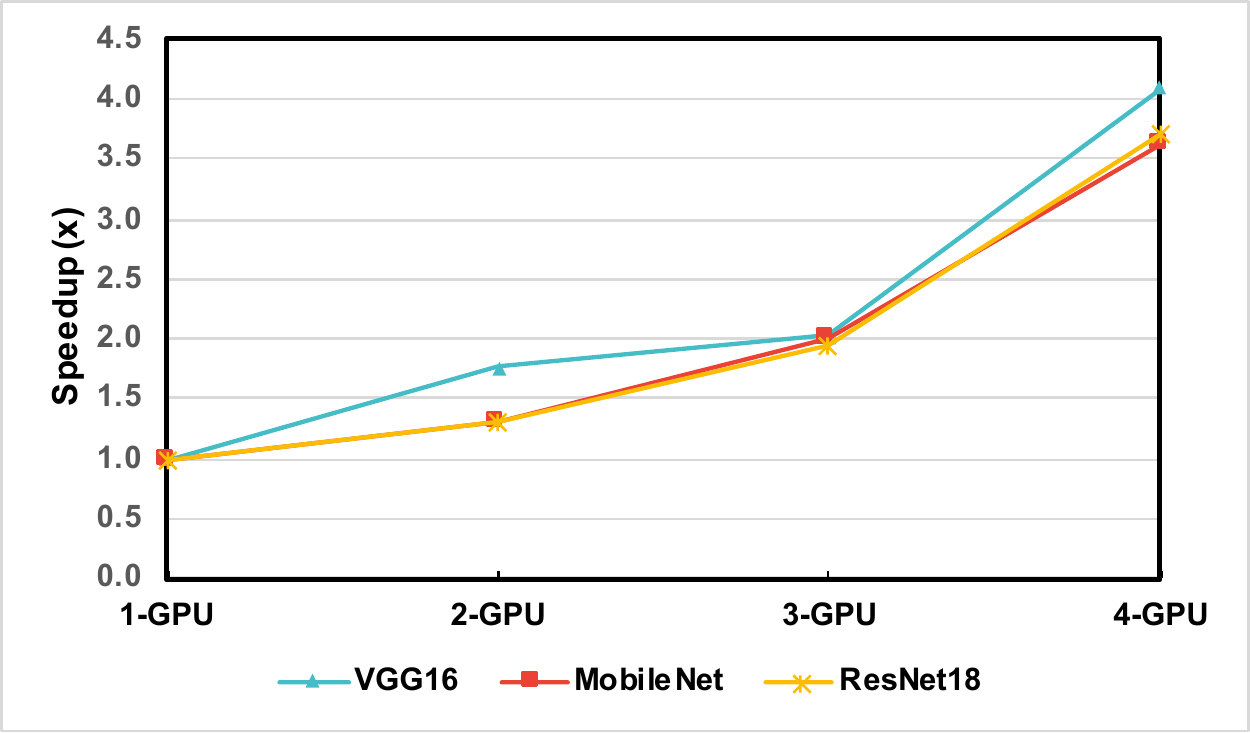}
    \caption{Multi-GPU scalability. \small{Note that runtime is normalized \textit{w.r.t} the performance for 1-GPU setting}.}
    \label{fig: Multi-GPU Scalability.}
    \vspace{-10pt}
\end{figure}
\section{Conclusion}
This paper introduces DSXplore, the first optimized design to explore the DSCs on CNNs.
Specifically, at the algorithm-level optimization, \Mname~incorporates a novel sliding-channel convolution (SCC), featured with the input-channel overlapping to capture cross-channel information that can effectively improve the accuracy while reducing FLOPs and parameter size across a board range of CNNs on mainstream image classification datasets. 
At the implementation level, we reduce the atomic operation during backward phase by leveraging the input-centric back-propagation design. 
Moreover, we fully integrated DSXplore with the Pytorch to improve programmability. Overall, our work paves a new way of exploring DSCs systematically and comprehensively through combining both algorithmic and implementation innovations.

\section{Acknowledgment}
This work was supported in part by NSF 1925717. 
Use was made of computational facilities purchased with funds from the National Science Foundation (OAC-1925717) and administered by the Center for Scientific Computing (CSC). The CSC is supported by the California NanoSystems Institute and the Materials Research Science and Engineering Center (MRSEC; NSF DMR 1720256) at UC Santa Barbara.

}

\bibliographystyle{IEEEtran}
\bibliography{reference}
\end{document}